\documentclass[conference]{article}
\usepackage{cite}
\usepackage{amsmath,amssymb,amsfonts}
\usepackage{algorithm}
\usepackage{algorithmic}
\usepackage{tabularx}
\usepackage{graphicx}
\usepackage{multirow}
\usepackage{subfig}
\usepackage{url}
\usepackage{textcomp}
\usepackage{amsmath}
\usepackage{rotating}
\usepackage{verbatim}
\usepackage{color}
\usepackage{changepage}
\usepackage{enumerate}
\usepackage{float}
\usepackage{authblk}
\usepackage{rotating}

\def\BibTeX{{\rm B\kern-.05em{\sc i\kern-.025em b}\kern-.08em
    T\kern-.1667em\lower.7ex\hbox{E}\kern-.125emX}}
\begin{document}

\title{A Comparison of Constraint Handling Techniques for Dynamic Constrained Optimization Problems
}
\author[1]{\rm Mar\'ia-Yaneli Ameca-Alducin}
\author[1]{\rm Maryam Hasani-Shoreh}
\author[1]{\rm Wilson Blaikie}
\author[1]{\rm Frank Neumann}
\author[2]{\rm Efr\'en Mezura-Montes}
\affil[1]{Optimisation and Logistics, School of Computer Science, The University of Adelaide SA 5005, Australia}
\affil[2]{Artificial Intelligence Research Center, University of Veracruz, Xalapa 91000, Mexico}
\affil[ ]{\textit {\{maria-yaneli.ameca-alducin, maryam.hasanishoreh\} @adelaide.edu.au, wilson.blaikie@student.adelaide.edu.au, frank.neumann@adelaide.edu.au, emezura@uv.mx}}





\maketitle

\begin{abstract}
Dynamic constrained optimization problems (DCOPs) have gained researchers attention in recent years because a vast majority of real world problems change over time. There are studies about the effect of constrained handling techniques in static optimization problems. However, there lacks any substantial study in the behavior of the most popular constraint handling techniques when dealing with DCOPs. In this paper we study the four most popular used constraint handling techniques and apply a simple Differential Evolution (DE) algorithm coupled with a change detection mechanism to observe the behavior of these techniques. These behaviors were analyzed using a common benchmark to determine which techniques are suitable for the most prevalent types of DCOPs. For the purpose of analysis, common measures in static environments were adapted to suit dynamic environments. While an overall superior technique could not be determined, certain techniques outperformed others in different aspects like rate of optimization or reliability of solutions.

\end{abstract}

{\bf Keywords:}
Dynamic Constrained Optimization; Constraint-Handling Techniques; Differential Evolution.
\section{Introduction}

Dynamic constrained optimization problems (DCOPs) have become very important in optimization research as many real world problems feature changing objective functions and/or constraints. 
Existing algorithms already find it difficult to optimize static constrained problems and it becomes even more difficult when constraints are dynamically changing~\cite{nguyen2012continuous}.
There currently exists a substantial amount of research into dynamic unconstrained optimization~\cite{Nguyen20121} and static constrained optimization~\cite{Mezura11} for evolutionary algorithms (EAs). However, this is not the case for dynamic constrained optimization.

One of the of the most important aspects of solving DCOPs is using an effective constraint handling technique to deal with the dynamic constraints in order to guide the search to those regions with feasible solutions and quickly adapt if constraints are changing.
In the specialized literature about DCOPs, the constraint handling techniques that have been applied include penalty function~\cite{CEC09}, repair methods~\cite{Das,bu2017continuous,nguyen2012continuous} and feasibility rules~\cite{ameca2014differential}. 

In a recent study, the impact of repair methods as a particular type of constraint handling techniques in DCOPs have been investigated~\cite{AmecaEvo2018}. While these methods show sound results for applying in DCOPs, other methods like $\epsilon$-constrained~\cite{takahama2005constrained} and stochastic ranking~\cite{runarsson2000stochastic} due to their characteristics seem to have competitive results in DCOPs. These characteristics mostly relate to the ability of the constraint handling method to increase or maintain diversity in the balance of feasible and infeasible solutions of the population. A comprehensive survey about the details of constraint handling techniques used with EAs can be found in~\cite{Mezura11}.
In $\epsilon$-constrained the infeasible solutions are treated more mildly compared to feasibility rules which implies that a higher diversity is usually  maintained. Similarly, stochastic ranking ranks the solutions not only based on the objective values and the feasibility of the solutions, but also a stochastic behavior is seen in the algorithm selection. This implies that infeasible solutions close to the region of feasibility are maintained in the population which may help when constraints change. 

In this paper we investigate stochastic ranking, $\epsilon$-constrained, penalty and feasibility rules as constraint handling techniques for dealing with DCOPs and compare these different approaches. In our comparison we do not consider repair methods because they are a mechanism that applies special operators to transform solutions~\cite{Mezura11} whereas the techniques being analyzed only manage solutions. Repair methods use extra evaluations during the optimization procedure compared to the constraint handling techniques~\cite{AmecaEvo2018}, this provides an unfair advantage in the results.
We investigate the strengths and weaknesses of these constraint handling techniques.  
Based on the offline error, feasibility and epsilon outperform the other techniques and maintain competitive performance with each other. However, the other techniques are more suited for alternative measures. Stochastic severely outperforms all other techniques in terms of speed, it makes up for its lack of reliability in how few evaluations it requires to find an optimum solution. While penalty is not the fastest nor does it have the least number of constraint violations, it is the most reliable of all the techniques and frequently returns the greatest number of successful solutions. Considering the proposed measure, the convergence score. Stochastic is also the highest performing technique for static constraints. However in the dynamic constraints, the techniques struggle to find successful solutions in the given time frame. A suggested solution to this issue is the addition of mechanisms to increase diversity or repair solutions to increase feasibility.

The remainder of this paper is organized as follows. In Section II the preliminaries are presented. An experimental design is introduced in Section III. The experimental analysis is conducted in Section IV. Finally the conclusion is drawn and future work is discussed in Section V.

\section{Preliminaries}
We now formally introduce Dynamic constrained optimization problems (DCOPs) and summaries the differential evolution algorithm and constraint handling methods that are subject to our investigations.
\subsection{Problem Statement}
Generally, a DCOP is considered as a kind of problem that its fitness function and feasible region will change by time \cite{DCOPS,Nguyen20121}. 
In mathematical terms, a DCOP is defined as follows: \\\\
Find $\vec{x}$, at each time $t$, which: 
\begin{equation}
		\min_{\vec{x}\in F_t\subseteq[L, U]} f(\vec{x}, t)
\end{equation}

\noindent where $t \in N^+$ is the current time, 
\begin{equation}
\begin{array}{l}
[L, U]=  \{ \vec{x} = (x_{1},x_{2},...,x_{D}) \mid L_i \leq x_i \leq U_i,\\
i = 1 \ldots D\}\\
\end{array}
\end{equation}
\noindent is the search space,

\noindent subject to:
\begin{equation}
\begin{array}{l}
F_{t}=\{ \vec{x} \mid \vec{x} \in [L,U], g_i (\vec{x},t) \le 0, i = 1, \ldots, m,\\
h_j (\vec{x},t) = 0,j = 1, \ldots, p\} \\
\end{array}
\end{equation}

\noindent is called the feasible region at time $t$.	

$\forall \vec{x} \in F_t$ if there exists a solution $\vec{x}^* \in F_t$ such that 
$f(\vec{x}^*,t)\leq f(\vec{x},t)$,  then $\vec{x}^*$ is called a feasible optimal solution 
and $f(\vec{x}^*,t)$ is called the feasible optima value at time $t$.

\subsection{Differential evolution with change detection mechanism}
Differential evolution (DE) was first introduced in \cite{ED1} as a stochastic search algorithm that is simple, reliable and fast. Each target vector in the population $\vec{x}_{i, G}$  generates one trial vector $\vec{u}_{i, G}$ by using a mutant vector $\vec{v}_{i,G}$. The mutation is applied through (\ref{eq:mutation}), where $\vec{x}_{r0,G}$, $\vec{x}_{r1,G}$, and $\vec{x}_{r2,G}$ are vectors chosen at random from the current population ($r0 \neq r1 \neq r2 \neq i$); $\vec{x}_{r0,G}$ is known as the base vector and $\vec{x}_{r1,G}$, and $\vec{x}_{r2,G}$ are the difference vectors and $F>0$ is a parameter called scale factor. 

\begin{equation}
\vec{v}_{i,G}= \vec{x}_{r0,G} + F (\vec{x}_{r1,G} - \vec{x}_{r2,G})
\label{eq:mutation}
\end{equation}

\noindent  The mutant vector $\vec{v}_{i, G}$ is then generated and combined with the target vector $\vec{x}_{i, G}$ to create the trial vector $\vec{u}_{i, G}$ by applying a crossover operator as shown in (\ref{eq:xover}).

\begin{equation}
	u_{i,j,G} = 
	\begin{cases} 
      v_{i,j,G}  & \mbox{if}(rand_{j} \leq CR) \mbox{ or } (j=J_{rand})   \\
      x_{i,j,G}  & \mbox{otherwise}
   \end{cases}
\label{eq:xover}
\end{equation}
 
\noindent where $CR \in [0,1]$ is the crossover probability, 
$rand_{j}$ generates a random real number which belongs to $[0,1]$, $j \in \left\lbrace 1,\ldots, D\right\rbrace$ is the $j$-th variable of the $D$-
dimensional vector, $J_{rand} \in [1,D]$ is a random integer which prevents it from choosing a target vector same as its trial vector. 

Overall, the best vector, based on its fitness function value, is selected  as for the next generation that is shown in (\ref{eq:replacement}) : 

\begin{equation}
	\vec{x}_{i,G+1} = 
	\begin{cases} 
      \vec{u}_{i,G}  & \mbox{if}(f(\vec{u}_{i,G}) \leq f(\vec{x}_{i,G})),\\
      \vec{x}_{i,G}  & \mbox{otherwise}
   \end{cases}
\label{eq:replacement}
\end{equation}

A general overview of this algorithm is presented in Algorithm \ref{alg:DE} and more details regarding to this Algorithm can be found in \cite{ED1, Mezura10a}.

\begin{equation}
\vec{v}_{i,G}= \vec{x}_{best,G} + F (\vec{x}_{r1,G} - \vec{x}_{r2,G})
\label{eq:mutation2}
\end{equation}

\begin{algorithm}[t]
\small
\begin{algorithmic}[1]
	 \STATE G=0
	 \STATE Create a randomly initial population $\vec{x}_{i,G}\,\forall i, i=1, \ldots, NP$
	 \STATE Evaluate $f(\vec{x}_{i,G})\,\forall i, i=1, \ldots, NP$
   \FOR{$G\gets 1$ to $MAX\_GEN$}
      \FOR{$i\gets 1$ to $NP$}
      			\IF{$i= 1$ or $i=NP/2$}
				\STATE Change Detection Mechanism ($\vec{x}_{i,G}$)
				\ENDIF
				\STATE Randomly select $r0 \neq r1 \neq r2 \neq i$
				\STATE $J_{rand} = randint [1,D]$
				\FOR{$j\gets 1$ to $D$}
					\IF{$rand_j \leq CR$ Or $j = J_{rand}$}
						\STATE $u_{i,j,G} = x_{r1,j,G} + F(x_{r2,j,G} - x_{r3,j,G}) $
					\ELSE
						\STATE $u_{i,j,G} = x_{i,j,G}$
					\ENDIF
				\ENDFOR
                \IF {$u_{i,j,G}$ is infeasible}
                \STATE {Use the repair method}
                \ENDIF
				\IF{$f(\vec{u}_{i,G}) \leq f(\vec{x}_{i,G})$}
					\STATE $\vec{x}_{i,G+1} = \vec{u}_{i,G}$
				\ELSE
					\STATE $\vec{x}_{i,G+1} = \vec{x}_{i,G}$
				\ENDIF
			\ENDFOR
			\STATE $G = G + 1$
		\ENDFOR
\end{algorithmic}
\caption{Differential Evolution Algorithm (DE/rand/1/bin)}
\label{alg:DE}
\end{algorithm}

For handling dynamism, in this paper a change detection mechanism is proposed that is based on calculating the error (see Equation~\ref{eq:error}) after each increase in the evaluations. This error is the difference between the values of the objective function and the optimum values at each time. In minimization problems the values of this error should be decreasing over generations. But if a change occurs this value may not be decreasing anymore.  
 
If any differences are detected, then all vectors in the current population are re-evaluated to get updated values.

\subsection{Constraint handling techniques}
\label{subsec:const handling tech}
The distinction between constraint handling techniques is the way they deal with the infeasible solutions. some of them like penalty function and feasibility rules are more strict about the infeasible solutions while others like $\epsilon$-constrained and stochastic ranking are more flexible with the in-feasibility of the solutions. The four constraint handling techniques are briefly reviewed in this section as follows. 

\subsubsection{Penalty}
The way that penalty works for handling constraints is that it tries to decrease the fitness of infeasible solutions in order to favor the selection of feasible solutions. There are different kinds of penalty methods including static (known as death), dynamic, adaptive, co-evolved and fuzzy-adapted. In this study we apply a simple version of penalty methods as follows. For each infeasible solution we use the following formula for objective function~\cite{RIGA}. 
\begin{equation}
f(\vec{x}, t)=f(\vec{x}, t)+2.5\phi(\vec{x},t)
\end{equation}
The sum of constraint violation $\phi(\vec{x},t)$ can be calculated as follows:
\begin{equation}
	\label{eq:sumcon}
	 \phi(\vec{x},t) =  \sum\limits_{i=1}^m max(0,g_i(\vec{x},t)) +  \sum\limits_{j=1}^p |h_i(\vec{x},t)|
\end{equation} 

where the values of each equality constraint $g_i (\vec{x},t), i = 1 \ldots m$ and also each equality constraint $h_j (\vec{x},t) = 0,j = 1 \ldots p$ are normalized.

\subsubsection{Feasibility rules}
One of the most popular constraint handling techniques used in bio-inspired algorithms is feasibility rules. This technique was proposed by Deb~\cite{deb2000efficient}, a set of three feasibility criteria are presented as follows:

\begin{enumerate}[i]
	\item Between 2 feasible vectors, the one with the highest fitness 
value is selected.
	\item If one vector is feasible and the other one is infeasible, the 
feasible vector is selected.
\item If both vectors are infeasible, the one with the lowest sum of 
constraint violation is selected.
\end{enumerate}

\subsubsection{$\epsilon$-constrained}
The $\epsilon$-constrained method was proposed by Takahama et al. in~\cite{takahama2005constrained}. This method is a type of transformation method that converts an algorithm for unconstrained optimization into an algorithm for constrained optimization. This technique has two main elements: 1) a relaxation of the limit to consider a solution as feasible, based on its sum of constraint violation previously defined in equation \ref{eq:sumcon}, with the aim of using its objective function value as a comparison criterion, and 2) a lexicographical ordering mechanism in which the minimization of the sum of constraint violation precedes the minimization of the objective function of a given problem. For any $\epsilon$ satisfying $\epsilon \geq 0$, the $\epsilon$ level comparisons $<_{\epsilon}$ and $\leq_{\epsilon}$ between $(f_1, \phi_1)$ and $(f_2, \phi_2)$ are defined in Equation~\ref{eq:caseless} and~\ref{eq:caseleq}.\\
\begin{figure*}
\begin{equation}
\label{eq:caseless}
    (f(\vec{x_1}), \phi(\vec{x_1}))  <_{\varepsilon}  (f(\vec{x_2}), \phi(\vec{x_2})) \Leftrightarrow
    \begin{cases}
            f(\vec{x_1}) < f(\vec{x_2}), & \text{if } \phi(\vec{x_1}), \phi(\vec{x_2}) \leq \varepsilon \\ 
            f(\vec{x_1}) < f(\vec{x_2}), & \text{if } \phi(\vec{x_1}) = \phi(\vec{x_2})  \\
            \phi(\vec{x_1}) < \phi(\vec{x_2}), & \text{otherwise} \\
    \end{cases}
\end{equation}

\begin{equation}
\label{eq:caseleq}
    (f(\vec{x_1}), \phi(\vec{x_1}))  \leq_{\varepsilon}  (f(\vec{x_2}), \phi(\vec{x_2})) \Leftrightarrow
            \begin {cases}
            f(\vec{x_1}) \leq f(\vec{x_2}), & \mbox{if } \phi(\vec{x_1}), \phi(\vec{x_2}) \leq \varepsilon \\ 
            f(\vec{x_1}) \leq f(\vec{x_2}), & \mbox{if } \phi(\vec{x_1}) = \phi(\vec{x_2})  \\
            \phi(\vec{x_1}) < \phi(\vec{x_2}), & \mbox{otherwise} \\
            \end{cases}
\end{equation}
\end{figure*}
When $\epsilon =0$,$<_0$ and $\leq_0$ are equivalent to the lexicographic order in which the constraint violation $\phi(\vec{x})$ precedes the function value $f(\vec{x})$. furthermore, in the case of $\epsilon=\infty$, the $\epsilon$ level comparisons $<$ and $\leq$ between function values.

\subsubsection{Stochastic ranking}
Runarsson and Yao proposed the stochastic ranking (SR) in~\cite{runarsson2000stochastic}. This technique was designed to deal with the shortcomings of a penalty function (that neither under- nor over-penalization is a good constraint handling technique and there should be a balance between preserving feasible individuals and rejecting infeasible ones). In SR, instead of the definition of penalty factors, a user-defined parameter called $P_f$ controls the criterion employed for comparison of infeasible solutions: 1) based on their sum of constraint violation or 2) based only on their objective function value. This technique uses a bubble-sort-like process to rank the solutions in the population, described in the algorithm \ref{alg:ranking}, where  $I$ is an individual of the population. $\phi(I_j)$ is the sum of constraint violation of individual $I_j$. $f(I_j)$ is the objective function value of individual $I_j$.

\begin{algorithm}[t]
\begin{algorithmic}[1]
        \FOR {$i = 1$ to $NP$}
            \FOR {$ j = 1$ to $NP-1$}
                \STATE u = random(0,1)
                \IF {($\phi(\vec{x_{j}},t) = \phi(\vec{x_{j+1}},t)) = 0$ \OR ($u < P_f$)} 
                    \IF {$f(\vec{x_{j}},t) > f(\vec{x_{j+1}},t) $}
                        \STATE Swap $\vec{x_{j}},t$ with  $\vec{x_{j+1}},t$
                    \ENDIF
                \ELSE
                    \IF {$\phi(\vec{x_{j}},t) > \phi(\vec{x_{j+1}},t)$}
                        \STATE Swap $\vec{x_{j}},t$ with  $\vec{x_{j+1}},t$
                    \ENDIF
                \ENDIF
            \ENDFOR
            \IF {swap not performed}
                \STATE break
            \ENDIF
        \ENDFOR
\end{algorithmic}
\caption{Stochastic Ranking sort algorithm ~\cite{runarsson2000stochastic}.}
\label{alg:ranking}
\end{algorithm} 

\section{Experimental Design}
For $\epsilon$-constrained method the value of $T_c$ is used in order to change the value of $\epsilon$ after a known amount of iterations.

\subsection{Test problems and performance measures}
\label{subsec:measures}
The chosen benchmark problem originally has 18 functions ~\cite{DCOPS}, however in this work, only 14 functions among them that are constrained were used for the experiments. 
The test problems in this benchmark consist of a variety of characteristics like i) disconnected feasible regions (1-3), ii) the global optima at the constraints' boundary or switchable between disconnected regions, or iii) the different shape and percentage of feasible area.
In the experiments, for the objective function, only medium severity is considered ($k=0.5$), while different change severities are considered for the constraints ($S=10$, $20$ and $50$). Based on the definition of the constrains in this benchmark ~\cite{DCOPS}, $S=10$ , $S=20$ and $S=50$ represent the severity of the changes on the constraints. The frequency of change ($f_c$) is considered equal to 1000 evaluations (only in the objective function).

For the purpose of comparing the effectiveness of each method, the following  performance measures were used:

\textbf{Modified offline error} ($M\_\text{off}\_e$)~\cite{nguyen2012continuous}: This measurement is equal to the average of the sum of errors in each generation divided by the total number of generations. Lower values for this measure is preferred and the zero value for offline error indicates a perfect performance~\cite{Nguyen20121}. This measure is defined in Equation~\ref{eq:offlineerror}.

\begin{equation}
\text{M\_off}\_e= \frac{1}{G_{max}} \sum_{G = 1}^{G_{max}} e(G)
\label{eq:offlineerror}
\end{equation}
\noindent where $G_{max}$ is the number of generations computed by the algorithm and $e(G)$ denotes the error 
in the current iteration $G$ (see~\ref{eq:error}):

\begin{equation}
	e(G)= |f(\vec{x}^*,t) - f(\vec{x}_{best,G},t)|
	\label{eq:error}
\end{equation}

\noindent where $f(\vec{x}^*,t)$ is the feasible global optima\footnote{This global optima is an approximation, which is the best solution found by DE in 30 runs for the current time.} at current time $t$, and $f(\vec{x}_{best,G},t)$ represent the best solution (feasible or infeasible) found so far at generation $G$  (for common offline error) at current time $t$. However for this modified version, in the case where the best solution is infeasible, the worst solution in the population is chosen instead of the best found. The reason for choosing the modified offline error was because in the common offline error, constraint violation is not considered, our main focus in the comparison of the constraint handling techniques is to know which one deals with the constraints more effectively. Without considering infeasible solutions the results were in favor of the methods that were more relaxing with the infeasible solutions.
We have applied other performance measures to observe other characteristics of these constraint handling techniques. These other measures are taken from proposed measures in ~\cite{mezura2012empirical} and we have modified them to be suitable for dynamic optimization. 

\textbf{Feasibility ratio ($FR_t$)}: The feasibility ratio consists on the number of feasible solutions per time ($f_t$) divided by the total number of times performed ($T$), as indicated in Equation ~\ref{eq:FR}.

\begin{equation}
FR_t=f_t/T
\label{eq:FR}
\end{equation}
The range of values for $FR_t$ goes from 0 to 1, where 1 means that in all times feasible solutions were found. In this way, a higher value is preferred.

\textbf{Success ratio ($SR_t$)}: The success ratio is calculated by the ratio of the number of successful times ($s_t$) \footnote{a time is considered successful if the best solution for this time is near to the optima with a precision ($10^{-4}$)} to the total number of times performed ($T$), as indicated in Equation~\ref{eq:SR}.

\begin{equation}
SR_t=s_t/T
\label{eq:SR}
\end{equation}
Similar to $FR_t$, the range of values for $SR_t$ goes from 0 to 1, where 1 means that in all of the times successful solutions were found. Therefore, a higher value is preferred.

\textbf{Average evaluations ($AE_t$)}: This measure is calculated by averaging the number of evaluations required on each successful run to find the first successful solution.
\begin{equation}
AE_t=(1/s_t) \cdot \sum_{i=1}^{s_t}(E_t) 
\end{equation}

where $E_t$ is the number of evaluations required to find the first successful solution in any successful time. For $E_t$, a lower value is preferred because it means that the average computational cost is lower for an algorithm to reach the vicinity of the feasible optimum solution.

\textbf{Convergence score ($CS_t$)}: The two previous performance measures ($SR_t$ and $AE_t$) are combined to measure the speed and reliability of an algorithm through a successful performance.
\begin{equation}
CS_t=AE_t/SR_t
\end{equation}

For this measure, a lower value is preferred because it means a better
ratio between speed and consistency of the algorithm.

\textbf{Progress ratio ($PR_t$)}: The objective is to measure
the improvement capability of the algorithm within the feasible region of the search space. For this measure high values are preferred because they indicate a higher improvement of the first feasible solution found. 

\begin{equation*}
PR_t=
\begin{cases}
 \bigl\lvert \ln\sqrt{\frac{f(\vec{x}_{first,G},t)}{f(\vec{x}_{best,G},t)}}\bigr\rvert& 
	\text{if } f(\vec{x}_{best,G},t) > 0  \\
 \bigl\lvert \ln\sqrt{\frac{f(\vec{x}_{first,G},t)+1}{f(\vec{x}_{best,G},t)+1}}\bigr\rvert& 
	\text{if } f(\vec{x}_{best,G},t) = 0  \text{  (18)}\\
 \bigl\lvert \ln\sqrt{\frac{f(\vec{x}_{first,G},t)+2|f(\vec{x}_{best,G},t)|}{f(\vec{x}_{best,G},t)+2|f(\vec{x}_{best,G},t)|}}\bigr\rvert& 
	\text{if } f(\vec{x}_{best,G},t) < 0  \\
	 \end{cases}
\end{equation*}

Where $f(\vec{x}_{first,G},t)$ is the value of the objective function of the first feasible solution found and $f(\vec{x}_{best,G},t)$ is the value of the objective function of the best solution found. For this measure, statistical values are also provided.

\begin{table*}[t]
\renewcommand{\arraystretch}{1.0}
\centering
\caption{\scriptsize Average and standard deviation of modified offline error values. Best results are remarked in boldface.}
\label{tab:offline values}
	\scalebox{0.55}{
\begin{tabular}{c|ccccccc}
\hline
\multirow{2}{*}{\textbf{Algorithms}} & \multicolumn{7}{c}{\textbf{$S=10$}}\\
 &&\textbf{G24\_3}&\textbf{G24\_3b}&\textbf{G24\_4}&\textbf{G24\_5}&\textbf{G24\_7}&\\\hline
Epsilon&&0.177($\pm$0.022)&0.23($\pm$0.026)&0.232($\pm$0.028)&0.223($\pm$0.031)&0.362($\pm$0.054)&\\
Feasibility&&\textbf{0.165($\pm$0.022})&\textbf{0.227($\pm$0.024})&\textbf{0.23($\pm$0.03})&\textbf{0.216($\pm$0.083})&\textbf{0.298($\pm$0.06})&\\
Penalty&&0.235($\pm$0.06)&0.491($\pm$0.225)&0.628($\pm$0.297)&2.316($\pm$1.521)&1.51($\pm$0.479)&\\
Stochastic&&0.219($\pm$0.047)&0.254($\pm$0.078)&0.231($\pm$0.057)&0.392($\pm$0.118)&0.457($\pm$0.124)&\\\hline
\multirow{2}{*}{\textbf{Algorithms}} & \multicolumn{7}{c}{\textbf{$S=20$}}\\
&\textbf{G24\_1}&\textbf{G24\_f}&\textbf{G24\_2}&\textbf{G24\_3}&\textbf{G24\_3b}&\textbf{G24\_3f}&\textbf{G24\_4}\\\hline
Epsilon&\textbf{0.25($\pm$0.04})&\textbf{0.028($\pm$0.011})&0.101($\pm$0.014)&0.179($\pm$0.037)&0.289($\pm$0.026)&\textbf{0.028($\pm$0.011})&0.282($\pm$0.039)\\
Feasibility&0.266($\pm$0.05)&0.032($\pm$0.019)&\textbf{0.097($\pm$0.017})&\textbf{0.148($\pm$0.015})&0.276($\pm$0.03)&0.077($\pm$0.258)&\textbf{0.273($\pm$0.033})\\
Penalty&0.714($\pm$0.392)&0.035($\pm$0.024)&1.142($\pm$0.977)&0.197($\pm$0.06)&0.706($\pm$0.297)&0.063($\pm$0.039)&0.729($\pm$0.357)\\
Stochastic&0.289($\pm$0.062)&0.227($\pm$0.143)&0.123($\pm$0.04)&0.203($\pm$0.044)&\textbf{0.258($\pm$0.046})&0.132($\pm$0.15)&0.277($\pm$0.056)\\\hline
 &\textbf{G24\_5}&\textbf{G24\_6a}&\textbf{G24\_6b}&\textbf{G24\_6c}&\textbf{G24\_6d}&\textbf{G24\_7}&\textbf{G24\_8b}\\\hline
Epsilon&0.158($\pm$0.022)&0.122($\pm$0.037)&0.087($\pm$0.012)&0.1($\pm$0.03)&0.143($\pm$0.04)&0.264($\pm$0.034)&0.285($\pm$0.039)\\
Feasibility&\textbf{0.141($\pm$0.017})&0.105($\pm$0.024)&\textbf{0.082($\pm$0.012})&\textbf{0.089($\pm$0.021})&0.169($\pm$0.062)&0.247($\pm$0.029)&\textbf{0.276($\pm$0.034})\\
Penalty&1.955($\pm$1.349)&0.247($\pm$0.079)&0.213($\pm$0.075)&0.284($\pm$0.077)&0.141($\pm$0.053)&0.704($\pm$0.153)&0.612($\pm$0.095)\\
Stochastic&0.162($\pm$0.04)&\textbf{0.091($\pm$0.022})&0.111($\pm$0.029)&0.103($\pm$0.027)&\textbf{0.138($\pm$0.039})&\textbf{0.204($\pm$0.053})&0.457($\pm$0.118)\\\hline
\multirow{2}{*}{\textbf{Algorithms}} & \multicolumn{7}{c}{\textbf{$S=50$}}\\
 &&\textbf{G24\_3}&\textbf{G24\_3b}&\textbf{G24\_4}&\textbf{G24\_5}&\textbf{G24\_7}&\\\hline
Epsilon&&0.174($\pm$0.036)&0.286($\pm$0.031)&0.282($\pm$0.035)&\textbf{0.13($\pm$0.021})&0.193($\pm$0.029)&\\
Feasibility&&\textbf{0.1($\pm$0.018})&0.257($\pm$0.05)&0.241($\pm$0.03)&0.135($\pm$0.023)&\textbf{0.188($\pm$0.03})&\\
Penalty&&0.122($\pm$0.04)&0.698($\pm$0.374)&0.718($\pm$0.382)&1.494($\pm$1.243)&0.385($\pm$0.09)&\\
Stochastic&&0.127($\pm$0.032)&\textbf{0.226($\pm$0.043})&\textbf{0.238($\pm$0.042})&0.146($\pm$0.045)&0.24($\pm$0.127)&\\\hline
\end{tabular}
}
\end{table*}

\section{Experimental analysis }

\begin{table} [t]
\centering
\caption{\scriptsize Statistical tests on the offline error values in Table 2.  ``X$^{(-)}$'' means that the corresponding algorithm outperformed algorithm X. ``X$^{(+)}$'' means that the corresponding algorithm was dominated by algorithm X. If algorithm X does not appear in column Y means no significant differences between X and Y.}
\label{tab:krus}
\scalebox{0.7}{

\begin{tabular}{l|cccc}
\hline

\multirow{2}{*}{\textbf{Functions}} & \multicolumn{4}{c}{\textbf{$S=10$}}\\
 &\textbf{Epsilon(1)}&\textbf{Feasibility(2)} &\textbf{Penalty(3)} &\textbf{Stochastic(4)}\\\hline
\textbf{G24\_3 (7.1-49.21\%)}&3$^{(-)}$, 4$^{(-)}$&3$^{(-)}$, 4$^{(-)}$ & 1$^{(+)}$, 2$^{(+)}$ & 1$^{(+)}$, 2$^{(+)}$ \\
\textbf{G24\_3b (7.1-49.21\%)}&3$^{(-)}$ &3$^{(-)}$  &1$^{(+)}$, 2$^{(+)}$, 4$^{(+)}$  &3$^{(-)}$ \\
\textbf{G24\_4 (0-44.2\%)}&3$^{(-)}$ &3$^{(-)}$  &1$^{(+)}$, 2$^{(+)}$, 4$^{(+)}$  &3$^{(-)}$ \\
\textbf{G24\_5 (0-44.2\%)}&3$^{(-)}$,  4$^{(-)}$ &  3$^{(-)}$, 4$^{(-)}$  &1$^{(+)}$, 2$^{(+)}$, 4$^{(+)}$&1$^{(+)}$, 2$^{(+)}$, 3$^{(-)}$ \\
\textbf{G24\_7 (0-44.2\%)}&3$^{(-)}$ &3$^{(-)}$, 4$^{(-)}$& 1$^{(+)}$, 2$^{(+)}$, 4$^{(+)}$& 2$^{(+)}$, 3$^{(-)}$\\
\hline
\multirow{2}{*}{\textbf{Functions}} & \multicolumn{4}{c}{\textbf{$S=20$}}\\
 &\textbf{Epsilon(1)}&\textbf{Feasibility(2)} &\textbf{Penalty(3)} &\textbf{Stochastic(4)}\\\hline
\textbf{G24\_1 (44.2\%)}&3$^{(-)}$ &3$^{(-)}$  &1$^{(+)}$, 2$^{(+)}$, 4$^{(+)}$  &3$^{(-)}$ \\
\textbf{G24\_f (44.2\%)}& 4$^{(-)}$ &4$^{(-)}$ &  4$^{(-)}$ & 1$^{(+)}$, 2$^{(+)}$, 3$^{(+)}$  \\
\textbf{G24\_2 (44.2\%)}&3$^{(-)}$ &3$^{(-)}$ &1$^{(+)}$, 2$^{(+)}$, 4$^{(+)}$  &3$^{(-)}$ \\
\textbf{G24\_3 (7.1-49.21\%)}&2$^{(+)}$&1$^{(-)}$, 3$^{(-)}$, 4$^{(-)}$&2$^{(+)}$ &2$^{(+)}$\\
\textbf{G24\_3b (7.1-49.21\%)}&3$^{(-)}$ &3$^{(-)}$  &1$^{(+)}$, 2$^{(+)}$, 4$^{(+)}$  &3$^{(-)}$ \\
\textbf{G24\_3f (7.1\%)}&3$^{(-)}$, 4$^{(-)}$ &3$^{(-)}$, 4$^{(-)}$ &1$^{(+)}$, 2$^{(+)}$& 1$^{(+)}$, 2$^{(+)}$ \\
\textbf{G24\_4 (0-44.2\%)}&3$^{(-)}$ &3$^{(-)}$  &1$^{(+)}$, 2$^{(+)}$, 4$^{(+)}$  &3$^{(-)}$ \\
\textbf{G24\_5 (0-44.2\%)}& 3$^{(-)}$& 3$^{(-)}$ &1$^{(+)}$, 2$^{(+)}$, 4$^{(+)}$ &3$^{(-)}$ \\
\textbf{G24\_6a (16.68\%)}&3$^{(-)}$, 4$^{(+)}$ &3$^{(-)}$  &1$^{(+)}$, 2$^{(+)}$, 4$^{(+)}$ &1$^{(-)}$, 3$^{(-)}$ \\
\textbf{G24\_6b (50.01\%)}&3$^{(-)}$, 4$^{(-)}$ &3$^{(-)}$, 4$^{(-)}$  &1$^{(+)}$, 2$^{(+)}$, 4$^{(+)}$ &1$^{(+)}$, 2$^{(+)}$,  3$^{(-)}$ \\
\textbf{G24\_6c (33.33\%)}&3$^{(-)}$ &3$^{(-)}$&1$^{(+)}$,2$^{(+)}$, 4$^{(+)}$&3$^{(-)}$ \\
\textbf{G24\_6d (20.91\%)}& - & 3$^{(+)}$, 4$^{(+)}$ & 2$^{(-)}$& 2$^{(-)}$\\
\textbf{G24\_7 (0-44.2\%)}&3$^{(-)}$, 4$^{(+)}$ &3$^{(-)}$, 4$^{(+)}$& 1$^{(+)}$, 2$^{(+)}$, 4$^{(+)}$& 1$^{(-)}$, 2$^{(-)}$, 3$^{(-)}$\\
\textbf{G24\_8b (44.2\%)}&3$^{(-)}$, 4$^{(-)}$ & 3$^{(-)}$, 4$^{(-)}$& 1$^{(+)}$, 2$^{(+)}$, 4$^{(+)}$ &1$^{(+)}$, 2$^{(+)}$, 3$^{(-)}$\\
\hline
\multirow{2}{*}{\textbf{Functions}} & \multicolumn{4}{c}{\textbf{$S=50$}}\\
 &\textbf{Epsilon(1)}&\textbf{Feasibility(2)} &\textbf{Penalty(3)} &\textbf{Stochastic(4)}\\\hline
\textbf{G24\_3 (7.1-49.21\%)}&2$^{(+)}$, 3$^{(+)}$, 4$^{(+)}$& 1$^{(-)}$, 4$^{(-)}$& 1$^{(-)}$ &1$^{(-)}$, 2$^{(+)}$\\
\textbf{G24\_3b (7.1-49.21\%)}&3$^{(-)}$, 4$^{(+)}$ &3$^{(-)}$ &1$^{(+)}$, 2$^{(+)}$, 4$^{(+)}$  &1$^{(-)}$, 3$^{(-)}$ \\
\textbf{G24\_4 (0-44.2\%)}&2$^{(+)}$, 3$^{(-)}$, 4$^{(+)}$& 1$^{(-)}$, 3$^{(-)}$ &1$^{(+)}$, 2$^{(+)}$, 4$^{(+)}$ &1$^{(-)}$, 3$^{(-)}$ \\
\textbf{G24\_5 (0-44.2\%)}& 3$^{(-)}$& 3$^{(-)}$ &1$^{(+)}$, 2$^{(+)}$, 4$^{(+)}$ &3$^{(-)}$ \\
\textbf{G24\_7 (0-44.2\%)}&3$^{(-)}$ &3$^{(-)}$& 1$^{(+)}$, 2$^{(+)}$, 4$^{(+)}$& 3$^{(-)}$\\
\hline

\end{tabular}
}
\end{table}
In the analysis, the effects of different severities on the constraints are considered for these fourteen test problems. We do not bring the results for changes of frequency since it does not have any effect in the behavior of the constraint handling techniques. 
The configurations for the experiments are as follows. The number of runs in the experiments are 30, and the number of considered times for dynamic perspective of the test algorithm is $5/k$ ($k=0.5$). 
Parameters relating to DE algorithm are as follows: DE variant is DE/rand/1/bin, population size is 20, scaling factor (F) is a random number $\in [0.2, 0.8]$, and crossover probability is 0.2. In the experiments, four constraint handling methods including $\epsilon$-constrained, feasibility rules, penalty function and stochastic ranking as explained in Section~\ref{subsec:const handling tech} have been applied for handling the constraint in DE algorithm.

\subsection{Experiment I: performance measure}

The results obtained for the four constraint handling techniques using modified offline error are summarized in Table~\ref{tab:offline values}. Furthermore, for the statistical validation, the 95\%-confidence Kruskal-Wallis test and the Bonferroni post hoc test, as suggested in~\cite{Derrac20113} are presented (see Table~\ref{tab:krus}). Non-parametric tests were adopted because the samples of runs did not fit to a normal distribution based on the Kolmogorov-Smirnov test.
Worth to mention that we removed the functions G24\_1, G24\_f, G24\_2, G24\_3f, G24\_6a, G24\_6b, G24\_6c, G24\_6d and G24\_8b from severity s=10 and 50 because they have static constraints. Therefore we include the results for these functions only for severity s=20 since they are the same for other severities as well. 

Table~\ref{tab:offline values}, illustrates the modified offline error values for different functions separated for each severity. From this table, one immediate conclusion is that penalty performed the worst among all techniques based on modified offline error values as it has higher error values for almost all of the functions, regardless of severity. However, to observe whether the methods have significant differences or not, the Kruskal-Wallis test has been carried out and the results are presented in Table~\ref{tab:krus}. The results of the statistical tests can be summarized as following observations:
in static constraint function G24\_6d penalty performed better than feasibility, and in function G24\_f it outperformed stochastic for severity s=20. In dynamic constraint function G24\_3, for severity s=50 it outperformed epsilon.

Among the techniques, epsilon and feasibility showed similar results. This is because epsilon uses a modification of feasibility rules, thus they have a similar trend to handling the constraints. This causes the two to lack significant difference in almost all of the functions excluding G24\_3 (for s=20 and 50) and G24\_4 (for s=50) where epsilon is the better performing technique.

In regards to stochastic, it was outperformed by both epsilon and feasibility in some functions like G24\_3, G24\_5 (s=10), G24\_f, G24\_3f, G24\_6b, G24\_8b (s=20) however, it only had significant difference with feasibility and not epsilon in functions G24\_7 (s=10) and G24\_3 (s=20 and 50). 

In general, severity did not have any significant effect on the results.
For testing other characteristics of the constraint handling techniques like feasibility probability, convergence rate, average number of function evaluations required for finding the first successful solution, convergence score and progress ratio to determine which performs the most effectively, other measures are defined and analyzed in the next section.

\subsection{Experiment II: behavior measures}

Tables~\ref{tab:allSeverity},~\ref{tab:allFunction} show the result of the measurements that were defined in Section~\ref{subsec:measures}. General observations regarding to the algorithms' behavior in these measures are summarized as follows.

Due to the lower rate of success ($SP_t$) in the stochastic ranking, this technique also tends to not find the optimum solution more often than its counterparts as shown in G24\_f and G24\_3f. This is attributed by the random nature of the stochastic ranking and its lack of consistent reliability as shown in all functions with non-zero success rates ($SP_t$).

Due to the large area of feasibility in this benchmark, the constraint handling techniques tended to have very high if not perfect feasibility rates ($FP_t$), however the penalty technique showed lower feasibility rates than its counterparts due to its nature of accepting infeasible solutions during optimization.

Based on the three measurements ($CS_t$, $AE_t$ and $SP_t$) in dynamic constraint functions including G24\_3, G24\_3b, G24\_4, G24\_7, with the exception of G24\_5, when the severity of the constraints is equal to 20 and 50, it is harder for the constraint handling techniques to converge with the optimal solutions. Conversely, for s=10, the constraint handling techniques are unable to converge to optimal solution for function G24\_7. Although this trend is also true for the static constraint function G24\_1.

For all of the functions, the three constraint handling techniques (epsilon, feasibility and penalty) had near identical success rates ($SP_t$), while not exactly the same they fell within one standard deviation of each other. However, the stochastic ranking technique had vastly different success rates compared to its counterparts.

Larger values for the progress ratio ($PR_t$) does not always indicate better performance since it depends on the distance between the first feasible solution and the best solution found. Even if the distance between these solutions is large, the best solution found can be stuck in a local optima and could never reach the global optimum. Indeed, the calculation of this measure does not take optimum values into consideration. 

Improvement of the constraint handling techniques would require additional optimization mechanisms as these techniques have very small standard deviation in the rate of optimization leading to similar progress ratio values.

\section{Conclusion and Future Work}
In this paper, we have compared common constraint handling techniques for solving DCOPs. For the measurements a modified version of offline error and other measures including average evaluations, convergence score, progress ratio,  feasibility ratio and successful ratio were adapted for dynamic environments and used for different severity of change of constraints. While the modified offline error data revealed competitive results between epsilon and feasibility, stochastic was considerably less reliable with large variations in the results and penalty presents the worst performance in terms of this measurement. However, stochastic managed the constraints and guided the algorithm to a successful solution much faster than any other technique albeit with a considerably lower reliability. This would make stochastic the more effective choice for simpler optimization problems where reliability is not an important factor in the performance. Conversely, penalty is the most reliable of the techniques which makes up for its lack of speed in constraint management, it takes far longer than the other techniques to reach a feasible solution but it consistently finds more successful solutions overall. Taking the proposed measure  (convergence score) into consideration, stochastic compensates for its unreliability with its speed and frequently scores the best out of the techniques in functions with static constraints. While this may be the case, in the functions with dynamic constraints, all of the techniques struggled to find successful solutions in the given time frame. This problem can be mitigated by adding additional mechanisms to the algorithms that increase its performance like methods of increasing diversity of solutions or repairing infeasible solutions.

In the future of dynamic constrained optimization, new constraint handling techniques would need to be developed to deal with the dynamic nature of the problem.

\section*{Acknowledgment}

This work has been supported through Australian Research Council (ARC) grant DP160102401.

\begin{sidewaystable}[h!] 
\centering
\caption{Average and standard deviation of average evaluations ($AE_t$), convergence score ($CS_t$), progress ratio ($PR_t$), feasibility ratio($FR_t$), successful ratio ($SR_t$). For s=10 and 50 only the functions that have dynamic constraints are displayed. Best results are remarked in boldface.}
\label{tab:allSeverity}
\scalebox{0.55}{
\hspace{-5em}
\begin{tabular}{ccccccccccccc}
\multicolumn{13}{c}{s = 10} \\ \hline
\multicolumn{1}{c|}{\multirow{2}{*}{Measures}} & \multicolumn{4}{c|}{G24\_3} & \multicolumn{4}{c|}{G24\_3b} & \multicolumn{4}{c}{G24\_4} \\ \cline{2-13} 
\multicolumn{1}{c|}{} & Epsilon & Feasibility & Penalty & \multicolumn{1}{c|}{Stochastic} & Epsilon & Feasibility & Penalty & \multicolumn{1}{c|}{Stochastic} & Epsilon & Feasibility & Penalty & Stochastic \\ \hline
\multicolumn{1}{c|}{$AE_t$} & 766.13$^{\pm(102.98)}$ & 709.17$^{\pm(164.94)}$ & 768.50$^{\pm(239.98)}$ & \multicolumn{1}{c|}{\textbf{329.33$^{\pm(244.60)}$} }
& 354.17$^{\pm(126.91)}$ & 340.17$^{\pm(128.77)}$ & 340.67$^{\pm(93.10)}$ & \multicolumn{1}{c|}{\textbf{145.23$^{\pm(114.67)}$ }}
& 382.10$^{\pm(128.28)}$ & 302.43$^{\pm(142.03)}$ & 330.70$^{\pm(102.17)}$ & \textbf{193.93$^{\pm(103.79)}$ }  \\
\multicolumn{1}{c|}{$CS_t$} & \textbf{3895.59$^{\pm(440.57)}$} & 4255.00$^{\pm(2006.64)}$ & 4116.96$^{\pm(3509.32)}$ & \multicolumn{1}{c|}{4490.91$^{\pm(901.67)}$ }
& 1264.88$^{\pm(400.47)}$ & 1308.33$^{\pm(493.97)}$ & 1148.31$^{\pm(400.45)}$ & \multicolumn{1}{c|}{\textbf{871.40$^{\pm(522.78)}$ }}
& 1432.88$^{\pm(434.23)}$ & 1226.08$^{\pm(480.36)}$ & 1127.39$^{\pm(533.76)}$ & \textbf{1077.41$^{\pm(531.86)}$}  \\
\multicolumn{1}{c|}{$PR_t$} & \textbf{1.09$^{\pm(0.00)}$} & \textbf{1.09$^{\pm(0.00)}$} & \textbf{1.09$^{\pm(0.00)}$} & \multicolumn{1}{c|}{1.07$^{\pm(0.02)}$ }
& \textbf{1.25$^{\pm(0.08)}$} & 1.21$^{\pm(0.05)}$ & 1.12$^{\pm(0.09)}$ & \multicolumn{1}{c|}{1.21$^{\pm(0.05)}$ }
& \textbf{1.23$^{\pm(0.06)}$} & 1.21$^{\pm(0.05)}$ & 1.15$^{\pm(0.07)}$ & 1.21$^{\pm(0.05)}$  \\
\multicolumn{1}{c|}{$FP_t$} & \textbf{0.83$^{\pm(0.00)}$} & \textbf{0.83$^{\pm(0.00)}$} & \textbf{0.83$^{\pm(0.00)}$} & \multicolumn{1}{c|}{0.82$^{\pm(0.01)}$ }
& \textbf{1.00$^{\pm(0.00)}$} & \textbf{1.00$^{\pm(0.00)}$} & 0.77$^{\pm(0.07)}$ & \multicolumn{1}{c|}{\textbf{1.00$^{\pm(0.00)}$} }
& \textbf{1.00$^{\pm(0.00)}$} & \textbf{1.00$^{\pm(0.00)}$} & 0.75$^{\pm(0.75)}$ & \textbf{1.00$^{\pm(0.00)}$}  \\
\multicolumn{1}{c|}{$SP_t$} & \textbf{0.20$^{\pm(0.02)}$} & 0.17$^{\pm(0.05)}$ & 0.19$^{\pm(0.03)}$ & \multicolumn{1}{c|}{0.07$^{\pm(0.06)}$ }
& 0.28$^{\pm(0.04)}$ & 0.26$^{\pm(0.06)}$ & \textbf{0.30$^{\pm(0.02)}$} & \multicolumn{1}{c|}{0.17$^{\pm(0.08)}$ } & 0.27$^{\pm(0.05)}$ & 0.25$^{\pm(0.05)}$ & \textbf{0.29$^{\pm(0.02)}$} & 0.18$^{\pm(0.06)}$ \\ \hline
\multicolumn{1}{c|}{\multirow{2}{*}{Measures}} & \multicolumn{4}{c|}{G24\_5} & \multicolumn{4}{c|}{G24\_7} &  &  &  &  \\ \cline{2-9}
\multicolumn{1}{c|}{} & Epsilon & Feasibility & Penalty & \multicolumn{1}{c|}{Stochastic} & Epsilon & Feasibility & Penalty & \multicolumn{1}{c|}{Stochastic} &  &  &  &  \\\cline{1-9}
\multicolumn{1}{c|}{$AE_t$} & 128.70$^{\pm(67.83)}$ & 124.63$^{\pm(67.32)}$ & 133.00$^{\pm(41.09)}$ & \multicolumn{1}{c|}{\textbf{49.50$^{\pm(56.80)}$ }}
& NaN & NaN & NaN & \multicolumn{1}{c|}{NaN} &  &  &  &  \\
\multicolumn{1}{c|}{$CS_t$} & 270.00$^{\pm(132.54)}$ & 261.47$^{\pm(132.13)}$ & 271.43$^{\pm(84.44)}$ & \multicolumn{1}{c|}{\textbf{126.92$^{\pm(226.41)}$ }} & NaN & NaN & NaN & \multicolumn{1}{c|}{NaN} &  &  &  &  \\
\multicolumn{1}{c|}{$PR_t$} & \textbf{0.43$^{\pm(0.05)}$} & 0.42$^{\pm(0.04)}$ & 0.26$^{\pm(0.03)}$ & \multicolumn{1}{c|}{0.40$^{\pm(0.07)}$ } & \textbf{0.59$^{\pm(0.02)}$} & 0.59$^{\pm(0.05)}$ & 0.50$^{\pm(0.03)}$ & \multicolumn{1}{c|}{0.55$^{\pm(0.03)}$} &  &  &  &  \\
\multicolumn{1}{c|}{$FP_t$} & 0.91$^{\pm(0.03)}$ & \textbf{0.91$^{\pm(0.02)}$} & 0.71$^{\pm(0.71)}$ & \multicolumn{1}{c|}{0.89$^{\pm(0.06)}$ }
& 0.92$^{\pm(0.04)}$ & \textbf{0.94$^{\pm(0.05)}$} & 0.59$^{\pm(0.59)}$ & \multicolumn{1}{c|}{0.89$^{\pm(0.09)}$} &  &  &  &  \\
\multicolumn{1}{c|}{$SP_t$} & 0.48$^{\pm(0.04)}$ & 0.48$^{\pm(0.04)}$ & \textbf{0.49$^{\pm(0.03)}$} & \multicolumn{1}{c|}{0.39$^{\pm(0.12)}$ }
& 0.00$^{\pm(0.00)}$ & 0.00$^{\pm(0.00)}$ & 0.00$^{\pm(0.00)}$ & \multicolumn{1}{c|}{0.00$^{\pm(0.00)}$} &  &  &  & \\\hline
\multicolumn{13}{c}{s = 50} \\ \hline
\multicolumn{1}{c|}{\multirow{2}{*}{Measures}} & \multicolumn{4}{c|}{G24\_3} & \multicolumn{4}{c|}{G24\_3b} & \multicolumn{4}{c}{G24\_4} \\ \cline{2-13} 
\multicolumn{1}{c|}{} & Epsilon & Feasibility & Penalty & \multicolumn{1}{c|}{Stochastic} & Epsilon & Feasibility & Penalty & \multicolumn{1}{c|}{Stochastic} & Epsilon & Feasibility & Penalty & Stochastic \\ \hline
\multicolumn{1}{c|}{$AE_t$} & NaN & NaN & NaN & \multicolumn{1}{c|}{NaN} & NaN & NaN & \textbf{31.47$^{\pm(0.00)}$} & \multicolumn{1}{c|}{NaN} & NaN & NaN & NaN & NaN \\
\multicolumn{1}{c|}{$CS_t$} & NaN & NaN & NaN & \multicolumn{1}{c|}{NaN} & NaN & NaN & \textbf{9440.00$^{\pm(0.00)}$} & \multicolumn{1}{c|}{NaN} & NaN & NaN & NaN & NaN \\
\multicolumn{1}{c|}{$PR_t$} & 0.88$^{\pm(0.00)}$ & 0.88$^{\pm(0.00)}$ & \textbf{0.89$^{\pm(0.01)}$} & \multicolumn{1}{c|}{0.87$^{\pm(0.01)}$ } & \textbf{0.98$^{\pm(0.06)}$} & 0.97$^{\pm(0.05)}$ & 0.92$^{\pm(0.14)}$ & \multicolumn{1}{c|}{0.98$^{\pm(0.09)}$ } & 0.96$^{\pm(0.07)}$ & 0.96$^{\pm(0.05)}$ & 0.88$^{\pm(0.13)}$ & \textbf{1.00$^{\pm(0.09)}$} \\
\multicolumn{1}{c|}{$FP_t$} & \textbf{1.00$^{\pm(0.00)}$} & \textbf{1.00$^{\pm(0.00)}$} & 0.67$^{\pm(0.12)}$ & \multicolumn{1}{c|}{\textbf{1.00$^{\pm(0.00)}$}} & \textbf{1.00$^{\pm(0.00)}$} & \textbf{1.00$^{\pm(0.00)}$} & 0.75$^{\pm(0.12)}$ & \multicolumn{1}{c|}{\textbf{1.00$^{\pm(0.00)}$}} & \textbf{1.00$^{\pm(0.00)}$} & \textbf{1.00$^{\pm(0.00)}$} & 0.70$^{\pm(0.17)}$ & \textbf{1.00$^{\pm(0.00)}$} \\
\multicolumn{1}{c|}{$SP_t$} & 0.00$^{\pm(0.00)}$ & 0.00$^{\pm(0.00)}$ & 0.00$^{\pm(0.00)}$ & \multicolumn{1}{c|}{0.00$^{\pm(0.00)}$} & 0.00$^{\pm(0.00)}$ & 0.00$^{\pm(0.00)}$ & \textbf{0.00$^{\pm(0.02)}$} & \multicolumn{1}{c|}{0.00$^{\pm(0.00)}$} & 0.00$^{\pm(0.00)}$ & 0.00$^{\pm(0.00)}$ & 0.00$^{\pm(0.00)}$ & 0.00$^{\pm(0.00)}$ \\ \hline
\multicolumn{1}{c|}{\multirow{2}{*}{Measures}} & \multicolumn{4}{c|}{G24\_5} & \multicolumn{4}{c|}{G24\_7} &  &  &  &  \\ \cline{2-9}
\multicolumn{1}{c|}{} & Epsilon & Feasibility & Penalty & \multicolumn{1}{c|}{Stochastic} & Epsilon & Feasibility & Penalty & \multicolumn{1}{c|}{Stochastic} &  &  &  &  \\ \cline{1-9}
\multicolumn{1}{c|}{$AE_t$} & 119.10$^{\pm(79.74)}$ & 126.10$^{\pm(70.38)}$ & 149.73$^{\pm(39.45)}$ & \multicolumn{1}{c|}{\textbf{34.13$^{\pm(38.16)}$} }
& NaN & NaN & NaN & \multicolumn{1}{c|}{NaN} &  &  &  &  \\
\multicolumn{1}{c|}{$CS_t$} & 255.21$^{\pm(157.67)}$ & 266.41$^{\pm(138.21)}$ & 303.51$^{\pm(89.38)}$ & \multicolumn{1}{c|}{\textbf{86.05$^{\pm(116.57)}$} } & NaN & NaN & NaN & \multicolumn{1}{c|}{NaN} &  &  &  &  \\
\multicolumn{1}{c|}{$PR_t$} & 0.70$^{\pm(0.04)}$ & \textbf{0.71$^{\pm(0.03)}$} & 0.70$^{\pm(0.09)}$ & \multicolumn{1}{c|}{0.71$^{\pm(0.06)}$ } & \textbf{0.97$^{\pm(0.00)}$} & 0.97$^{\pm(0.01)}$ & 0.96$^{\pm(0.02)}$ & \multicolumn{1}{c|}{0.95$^{\pm(0.02)}$} &  &  &  &  \\
\multicolumn{1}{c|}{$FP_t$} & \textbf{1.00$^{\pm(0.00)}$} & \textbf{1.00$^{\pm(0.00)}$} & 0.87$^{\pm(0.09)}$ & \multicolumn{1}{c|}{\textbf{1.00$^{\pm(0.00)}$}} & \textbf{1.00$^{\pm(0.00)}$} & \textbf{1.00$^{\pm(0.00)}$} & 0.74$^{\pm(0.14)}$ & \multicolumn{1}{c|}{\textbf{1.00$^{\pm(0.00)}$}} &  &  &  &  \\
\multicolumn{1}{c|}{$SP_t$} & 0.47$^{\pm(0.05)}$ & 0.47$^{\pm(0.04)}$ & \textbf{0.49$^{\pm(0.02)}$} & \multicolumn{1}{c|}{0.40$^{\pm(0.08)}$ } & 0.00$^{\pm(0.00)}$ & 0.00$^{\pm(0.00)}$ & 0.00$^{\pm(0.00)}$ & \multicolumn{1}{c|}{0.00$^{\pm(0.00)}$} &  &  &  & 
\end{tabular}
}
\end{sidewaystable}

\begin{sidewaystable}[h!] 
\centering
\caption{Average and standard deviation of average evaluations ($AE_t$), convergence score ($CS_t$), progress ratio ($PR_t$), feasibility ratio($FR_t$), successful ratio ($SR_t$). For s=20 all the 14 functions are displayed. Best results are remarked in boldface.}
\label{tab:allFunction}
\scalebox{0.55}{
\hspace{-5em}

\begin{tabular}{ccccccccccccc}
\multicolumn{13}{c}{s = 20} \\ \hline
\multicolumn{1}{c|}{\multirow{2}{*}{Measures}} & \multicolumn{4}{c|}{G24\_1} & \multicolumn{4}{c|}{G24\_f} & \multicolumn{4}{c}{G24\_2} \\ \cline{2-13} 
\multicolumn{1}{c|}{} & Epsilon & Feasibility & Penalty & \multicolumn{1}{c|}{Stochastic} & Epsilon & Feasibility & Penalty & \multicolumn{1}{c|}{Stochastic} & Epsilon & Feasibility & Penalty & Stochastic \\ \hline
\multicolumn{1}{c|}{$AE_t$} & NaN & NaN & NaN & \multicolumn{1}{c|}{NaN}
& 72.30$^{\pm(35.80)}$ & \textbf{66.17$^{\pm(39.24)}$} & 126.30$^{\pm(76.77)}$ & \multicolumn{1}{c|}{NaN}
& 47.50$^{\pm(30.33)}$ & 41.90$^{\pm(20.01)}$ & 41.07$^{\pm(26.55)}$ & \textbf{33.70$^{\pm(20.21)}$} \\
\multicolumn{1}{c|}{$CS_t$} & NaN & NaN & NaN & \multicolumn{1}{c|}{NaN}
& 98.14$^{\pm(45.15)}$ & \textbf{89.41$^{\pm(51.01)}$} & 184.83$^{\pm(155.89)}$ & \multicolumn{1}{c|}{NaN}
& 79.17$^{\pm(50.55)}$ & 69.83$^{\pm(33.34)}$ & 68.44$^{\pm(44.25)}$ & \textbf{68.31$^{\pm(37.12)}$} \\
\multicolumn{1}{c|}{$PR_t$} & \textbf{1.12$^{\pm(0.01)}$} & \textbf{1.12$^{\pm(0.01)}$} & 1.01$^{\pm(0.16)}$ & \multicolumn{1}{c|}{1.09$^{\pm(0.02)}$ }
& \textbf{1.03$^{\pm(0.00)}$} & \textbf{1.03$^{\pm(0.00)}$} & \textbf{1.03$^{\pm(0.00)}$} & \multicolumn{1}{c|}{1.01$^{\pm(0.02)}$ }
& \textbf{0.59$^{\pm(0.00)}$} & \textbf{0.59$^{\pm(0.00)}$} & 0.57$^{\pm(0.04)}$ & 0.57$^{\pm(0.01)}$ \\
\multicolumn{1}{c|}{$FP_t$} & \textbf{1.00$^{\pm(0.00)}$} & \textbf{1.00$^{\pm(0.00)}$} & 0.72$^{\pm(0.15)}$ & \multicolumn{1}{c|}{\textbf{1.00$^{\pm(0.00)}$} }
& \textbf{1.00$^{\pm(0.00)}$} & \textbf{1.00$^{\pm(0.00)}$} & 0.79$^{\pm(0.13)}$ & \multicolumn{1}{c|}{\textbf{1.00$^{\pm(0.00)}$} }
& \textbf{1.00$^{\pm(0.00)}$} & \textbf{1.00$^{\pm(0.00)}$} & 0.89$^{\pm(0.11)}$ & \textbf{1.00$^{\pm(0.00)}$} \\
\multicolumn{1}{c|}{$SP_t$} & 0.00$^{\pm(0.00)}$ & 0.00$^{\pm(0.00)}$ & 0.00$^{\pm(0.00)}$ & \multicolumn{1}{c|}{0.00$^{\pm(0.00)}$}
& \textbf{0.74$^{\pm(0.05)}$} & 0.74$^{\pm(0.06)}$ & 0.68$^{\pm(0.12)}$ & \multicolumn{1}{c|}{0.00$^{\pm(0.00)}$ }
& \textbf{0.60$^{\pm(0.00)}$} & \textbf{0.60$^{\pm(0.00)}$} & \textbf{0.60$^{\pm(0.00)}$} & 0.49$^{\pm(0.09)}$ \\ \hline
\multicolumn{1}{c|}{\multirow{2}{*}{Measures}} & \multicolumn{4}{c|}{G24\_3} & \multicolumn{4}{c|}{G24\_3b} & \multicolumn{4}{c}{G24\_3f} \\ \cline{2-13} 
\multicolumn{1}{c|}{} & Epsilon & Feasibility & Penalty & \multicolumn{1}{c|}{Stochastic} & Epsilon & Feasibility & Penalty & \multicolumn{1}{c|}{Stochastic} & Epsilon & Feasibility & Penalty & Stochastic \\ \hline
\multicolumn{1}{c|}{$AE_t$} & NaN & NaN & NaN & \multicolumn{1}{c|}{NaN}
& NaN & NaN & NaN & \multicolumn{1}{c|}{NaN}
& 72.73$^{\pm(36.65)}$ & \textbf{58.50$^{\pm(40.80)}$} & 93.47$^{\pm(56.75)}$ & NaN \\
\multicolumn{1}{c|}{$CS_t$} & NaN & NaN & NaN & \multicolumn{1}{c|}{NaN}
& NaN & NaN & NaN & \multicolumn{1}{c|}{NaN}
& 98.73$^{\pm(46.59)}$ & \textbf{83.57$^{\pm(51.31)}$} & 136.78$^{\pm(95.69)}$ & NaN \\
\multicolumn{1}{c|}{$PR_t$} & 0.96$^{\pm(0.01)}$ & 0.96$^{\pm(0.00)}$ & \textbf{0.97$^{\pm(0.01)}$} & \multicolumn{1}{c|}{0.95$^{\pm(0.01)}$ }
& \textbf{1.17$^{\pm(0.09)}$} & 1.16$^{\pm(0.08)}$ & 0.97$^{\pm(0.16)}$ & \multicolumn{1}{c|}{1.12$^{\pm(0.09)}$}
& \textbf{0.83$^{\pm(0.00)}$} & 0.82$^{\pm(0.03)}$ & \textbf{0.83$^{\pm(0.00)}$} & 0.82$^{\pm(0.01)}$ \\
\multicolumn{1}{c|}{$FP_t$} & \textbf{1.00$^{\pm(0.00)}$} & \textbf{1.00$^{\pm(0.00)}$} & 0.72$^{\pm(0.14)}$ & \multicolumn{1}{c|}{\textbf{1.00$^{\pm(0.00)}$} }
& \textbf{1.00$^{\pm(0.00)}$} & \textbf{1.00$^{\pm(0.00)}$} & 0.66$^{\pm(0.19)}$ & \multicolumn{1}{c|}{\textbf{1.00$^{\pm(0.00)}$} }
& \textbf{1.00$^{\pm(0.00)}$} & \textbf{1.00$^{\pm(0.00)}$} & 0.78$^{\pm(0.10)}$ & \textbf{1.00$^{\pm(0.00)}$} \\
\multicolumn{1}{c|}{$SP_t$} & 0.00$^{\pm(0.00)}$ & 0.00$^{\pm(0.00)}$ & 0.00$^{\pm(0.00)}$ & \multicolumn{1}{c|}{0.00$^{\pm(0.00)}$}
& 0.00$^{\pm(0.00)}$ & 0.00$^{\pm(0.00)}$ & 0.00$^{\pm(0.00)}$ & \multicolumn{1}{c|}{0.00$^{\pm(0.00)}$}
& \textbf{0.74$^{\pm(0.05)}$} & 0.70$^{\pm(0.14)}$ & 0.68$^{\pm(0.09)}$ & 0.00$^{\pm(0.00)}$\\\hline
\multicolumn{1}{c|}{\multirow{2}{*}{Measures}} & \multicolumn{4}{c|}{G24\_4} & \multicolumn{4}{c|}{G24\_5} & \multicolumn{4}{c}{G24\_6a} \\ \cline{2-13} 
\multicolumn{1}{c|}{} & Epsilon & Feasibility & Penalty & \multicolumn{1}{c|}{Stochastic} & Epsilon & Feasibility & Penalty & \multicolumn{1}{c|}{Stochastic} & Epsilon & Feasibility & Penalty & Stochastic \\ \hline
\multicolumn{1}{c|}{$AE_t$} & NaN & NaN & NaN & \multicolumn{1}{c|}{NaN}
& 124.47$^{\pm(67.67)}$ & 126.37$^{\pm(57.62)}$ & 148.83$^{\pm(42.22)}$ & \multicolumn{1}{c|}{\textbf{23.37$^{\pm(22.30)}$} }
& 793.57$^{\pm(47.74)}$ & 793.70$^{\pm(54.25)}$ & 747.50$^{\pm(95.12)}$ & \textbf{556.90$^{\pm(103.85)}$} \\
\multicolumn{1}{c|}{$CS_t$} & NaN & NaN & NaN & \multicolumn{1}{c|}{NaN}
& 261.12$^{\pm(132.88)}$ & 263.26$^{\pm(112.28)}$ & 303.74$^{\pm(95.67)}$ & \multicolumn{1}{c|}{\textbf{57.46$^{\pm(44.34)}$} }
& 1214.64$^{\pm(398.88)}$ & 1253.21$^{\pm(434.65)}$ & \textbf{942.23$^{\pm(294.79)}$} & 1622.04$^{\pm(1157.78)}$ \\
\multicolumn{1}{c|}{$PR_t$} & \textbf{1.15$^{\pm(0.06)}$} & 1.15$^{\pm(0.09)}$ & 1.01$^{\pm(0.12)}$ & \multicolumn{1}{c|}{1.11$^{\pm(0.08)}$ }
& \textbf{0.62$^{\pm(0.02)}$} & \textbf{0.62$^{\pm(0.02)}$} & 0.60$^{\pm(0.06)}$ & \multicolumn{1}{c|}{0.61$^{\pm(0.03)}$ }
& 0.91$^{\pm(0.00)}$ & 0.91$^{\pm(0.00)}$ & 0.85$^{\pm(0.11)}$ & \textbf{0.92$^{\pm(0.01)}$} \\
\multicolumn{1}{c|}{$FP_t$} & \textbf{1.00$^{\pm(0.00)}$} & \textbf{1.00$^{\pm(0.00)}$} & 0.71$^{\pm(0.16)}$ & \multicolumn{1}{c|}{\textbf{1.00$^{\pm(0.00)}$} }
& \textbf{1.00$^{\pm(0.00)}$} & \textbf{1.00$^{\pm(0.00)}$} & 0.84$^{\pm(0.10)}$ & \multicolumn{1}{c|}{\textbf{1.00$^{\pm(0.00)}$} }
& \textbf{1.00$^{\pm(0.00)}$} & \textbf{1.00$^{\pm(0.00)}$} & 0.94$^{\pm(0.09)}$ & \textbf{1.00$^{\pm(0.00)}$} \\
\multicolumn{1}{c|}{$SP_t$} & 0.00$^{\pm(0.00)}$ & 0.00$^{\pm(0.00)}$ & 0.00$^{\pm(0.00)}$ & \multicolumn{1}{c|}{0.00$^{\pm(0.00)}$}
& 0.48$^{\pm(0.04)}$ & 0.48$^{\pm(0.04)}$ & \textbf{0.49$^{\pm(0.03)}$} & \multicolumn{1}{c|}{0.41$^{\pm(0.02)}$ }
& 0.65$^{\pm(0.14)}$ & 0.63$^{\pm(0.16)}$ & \textbf{0.79$^{\pm(0.12)}$} & 0.34$^{\pm(0.12)}$ \\ \hline
\multicolumn{1}{c|}{\multirow{2}{*}{Measures}} & \multicolumn{4}{c|}{G24\_6b} & \multicolumn{4}{c|}{G24\_6c} & \multicolumn{4}{c}{G24\_6d} \\ \cline{2-13} 
\multicolumn{1}{c|}{} & Epsilon & Feasibility & Penalty & \multicolumn{1}{c|}{Stochastic} & Epsilon & Feasibility & Penalty & \multicolumn{1}{c|}{Stochastic} & Epsilon & Feasibility & Penalty & Stochastic \\ \hline
\multicolumn{1}{c|}{$AE_t$} & 730.27$^{\pm(62.03)}$ & 712.43$^{\pm(57.65)}$ & 748.07$^{\pm(112.38)}$ & \multicolumn{1}{c|}{\textbf{506.27$^{\pm(109.47)}$}}
& 717.53$^{\pm(49.96)}$ & 719.23$^{\pm(45.86)}$ & 789.23$^{\pm(147.43)}$ & \multicolumn{1}{c|}{\textbf{505.27$^{\pm(123.24)}$} }
& 646.50$^{\pm(45.52)}$ & 656.73$^{\pm(42.52)}$ & 626.10$^{\pm(40.45)}$ & \textbf{421.33$^{\pm(71.61)}$} \\
\multicolumn{1}{c|}{$CS_t$} & 1023.74$^{\pm(152.18)}$ & \textbf{984.93$^{\pm(93.95)}$} & 1133.43$^{\pm(465.92)}$ & \multicolumn{1}{c|}{1393.39$^{\pm(1035.71)}$}
& 956.71$^{\pm(141.80)}$ & \textbf{942.23$^{\pm(175.47)}$} & 1143.82$^{\pm(497.61)}$ & \multicolumn{1}{c|}{1471.65$^{\pm(1106.47)}$ }
& \textbf{682.92$^{\pm(72.61)}$} & 772.63$^{\pm(133.90)}$ & 688.02$^{\pm(99.53)}$ & 1181.31$^{\pm(647.70)}$ \\
\multicolumn{1}{c|}{$PR_t$} & 0.91$^{\pm(0.00)}$ & 0.91$^{\pm(0.00)}$ & 0.83$^{\pm(0.12)}$ & \multicolumn{1}{c|}{\textbf{0.92$^{\pm(0.01)}$}}
& 0.91$^{\pm(0.00)}$ & 0.91$^{\pm(0.02)}$ & 0.88$^{\pm(0.05)}$ & \multicolumn{1}{c|}{\textbf{0.92$^{\pm(0.01)}$} }
& 1.41$^{\pm(0.07)}$ & 1.42$^{\pm(0.10)}$ & \textbf{1.43$^{\pm(0.10)}$} & 1.40$^{\pm(0.08)}$ \\
\multicolumn{1}{c|}{$FP_t$} & \textbf{1.00$^{\pm(0.00)}$} & \textbf{1.00$^{\pm(0.00)}$} & 0.85$^{\pm(0.10)}$ & \multicolumn{1}{c|}{\textbf{1.00$^{\pm(0.00)}$} }
& \textbf{1.00$^{\pm(0.00)}$} & \textbf{1.00$^{\pm(0.00)}$} & 0.86$^{\pm(0.10)}$ & \multicolumn{1}{c|}{\textbf{1.00$^{\pm(0.00)}$} }
& \textbf{1.00$^{\pm(0.00)}$} & \textbf{1.00$^{\pm(0.00)}$} & \textbf{1.00$^{\pm(0.00)}$} & \textbf{1.00$^{\pm(0.00)}$} \\
\multicolumn{1}{c|}{$SP_t$} & 0.71$^{\pm(0.10)}$ & \textbf{0.72$^{\pm(0.09)}$} & 0.66$^{\pm(0.11)}$ & \multicolumn{1}{c|}{0.36$^{\pm(0.18)}$}
& 0.75$^{\pm(0.11)}$ & \textbf{0.76$^{\pm(0.12)}$} & 0.69$^{\pm(0.13)}$ & \multicolumn{1}{c|}{0.34$^{\pm(0.15)}$ }
& \textbf{0.95$^{\pm(0.07)}$} & 0.85$^{\pm(0.13)}$ & 0.91$^{\pm(0.10)}$ & 0.36$^{\pm(0.14)}$ \\\hline
\multicolumn{1}{c|}{\multirow{2}{*}{Measures}} & \multicolumn{4}{c|}{G24\_7} & \multicolumn{4}{c|}{G24\_8b} & \multicolumn{4}{c}{} \\ \cline{2-9}  
\multicolumn{1}{c|}{} & Epsilon & Feasibility & Penalty & \multicolumn{1}{c|}{Stochastic} & Epsilon & Feasibility & Penalty & \multicolumn{1}{c|}{Stochastic} &  &  &  &  \\\cline{1-9}
\multicolumn{1}{c|}{$AE_t$} & NaN & NaN & NaN & \multicolumn{1}{c|}{NaN}
& 242.37$^{\pm(404.83)}$ & 260.27$^{\pm(401.82)}$ & 348.40$^{\pm(441.93)}$ & \multicolumn{1}{c|}{\textbf{84.70$^{\pm(221.61)}$} }
&  &  &  &  \\
\multicolumn{1}{c|}{$CS_t$} & NaN & NaN & NaN & \multicolumn{1}{c|}{NaN}
& 9088.75$^{\pm(484.85)}$ & 7808.00$^{\pm(930.57)}$ & 7488.57$^{\pm(1471.50)}$ & \multicolumn{1}{c|}{\textbf{6352.50$^{\pm(497.94)}$} }
&  &  &  &  \\
\multicolumn{1}{c|}{$PR_t$} & 0.88$^{\pm(0.01)}$ & \textbf{0.88$^{\pm(0.00)}$} & 0.85$^{\pm(0.03)}$ & \multicolumn{1}{c|}{0.88$^{\pm(0.01)}$ }
& \textbf{0.94$^{\pm(0.05)}$} & \textbf{0.94$^{\pm(0.05)}$} & 0.64$^{\pm(0.19)}$ & \multicolumn{1}{c|}{0.77$^{\pm(0.07)}$ }
&  &  &  &  \\
\multicolumn{1}{c|}{$FP_t$} & \textbf{1.00$^{\pm(0.00)}$} & \textbf{1.00$^{\pm(0.00)}$} & 0.66$^{\pm(0.15)}$ & \multicolumn{1}{c|}{\textbf{1.00$^{\pm(0.00)}$} }
& \textbf{1.00$^{\pm(0.00)}$} & \textbf{1.00$^{\pm(0.00)}$} & 0.57$^{\pm(0.24)}$ & \multicolumn{1}{c|}{\textbf{1.00$^{\pm(0.00)}$} } &  &  &  &  \\
\multicolumn{1}{c|}{$SP_t$} & 0.00$^{\pm(0.00)}$ & 0.00$^{\pm(0.00)}$ & 0.00$^{\pm(0.00)}$ & \multicolumn{1}{c|}{0.00$^{\pm(0.00)}$}
& 0.03$^{\pm(0.04)}$ & 0.03$^{\pm(0.05)}$ & \textbf{0.14$^{\pm(0.10)}$} & \multicolumn{1}{c|}{0.01$^{\pm(0.03)}$ }
&  &  &  &  \\
\end{tabular}

}
\end{sidewaystable} 

\begin{thebibliography}{9}
\bibitem{nguyen2012continuous}
T.~T. Nguyen and X.~Yao, ``Continuous dynamic constrained optimization—the
  challenges,'' \emph{IEEE Transactions on Evolutionary Computation}, vol.~16,
  no.~6, pp. 769--786, 2012.

\bibitem{Nguyen20121}
T.~Nguyen, S.~Yang, and J.~Branke, ``Evolutionary dynamic optimization: A
  survey of the state of the art,'' \emph{Swarm and Evolutionary Computation},
  vol.~6, no.~0, pp. 1 -- 24, 2012. [Online]. Available:
  \url{http://www.sciencedirect.com/science/article/pii/S2210650212000363}

\bibitem{Mezura11}
E.~Mezura-Montes and C.~A.~C. Coello, ``Constraint-handling in nature-inspired
  numerical optimization: Past, present and future,'' \emph{Swarm and
  Evolutionary Computation}, vol.~1, no.~4, pp. 173--194, 2011.

\bibitem{CEC09}
T.~T. Nguyen and X.~Yao, ``Benchmarking and solving dynamic constrained
  problems,'' in \emph{Evolutionary Computation, 2009. CEC '09. IEEE Congress
  on}, 2009, pp. 690--697.

\bibitem{Das}
K.~Pal, C.~Saha, S.~Das, and C.~Coello-Coello, ``Dynamic constrained
  optimization with offspring repair based gravitational search algorithm,'' in
  \emph{Evolutionary Computation (CEC), 2013 IEEE Congress on}, 2013, pp.
  2414--2421.

\bibitem{bu2017continuous}
C.~Bu, W.~Luo, and L.~Yue, ``Continuous dynamic constrained optimization with
  ensemble of locating and tracking feasible regions strategies,'' \emph{IEEE
  Transactions on Evolutionary Computation}, vol.~21, no.~1, pp. 14--33, 2017.

\bibitem{ameca2014differential}
M.-Y. Ameca-Alducin, E.~Mezura-Montes, and N.~Cruz-Ramirez, ``Differential
  evolution with combined variants for dynamic constrained optimization,'' in
  \emph{Evolutionary computation (CEC), 2014 IEEE congress on}.\hskip 1em plus
  0.5em minus 0.4em\relax IEEE, 2014, pp. 975--982.

\bibitem{AmecaEvo2018}
M.-Y. Ameca-Alducin, M.~Hasani-Shoreh, and F.~Neumann, ``On the use of repair
  methods in differential evolution for dynamic constrained optimization,'' in
  \emph{Applications of Evolutionary Computation}, P.~Kaufmann and K.~Sim,
  Eds.\hskip 1em plus 0.5em minus 0.4em\relax Cham: Springer International
  Publishing, 2018, pp.~--, in Press.

\bibitem{takahama2005constrained}
T.~Takahama, S.~Sakai, and N.~Iwane, ``Constrained optimization by the epsilon
  constrained hybrid algorithm of particle swarm optimization and genetic
  algorithm,'' in \emph{Australian Conference on Artificial Intelligence}, vol.
  3809.\hskip 1em plus 0.5em minus 0.4em\relax Springer, 2005, pp. 389--400.

\bibitem{runarsson2000stochastic}
T.~P. Runarsson and X.~Yao, ``Stochastic ranking for constrained evolutionary
  optimization,'' \emph{IEEE Transactions on evolutionary computation}, vol.~4,
  no.~3, pp. 284--294, 2000.

\bibitem{DCOPS}
T.~Nguyen and X.~Yao, ``Continuous dynamic constrained optimization: The
  challenges,'' \emph{IEEE Transactions on Evolutionary Computation}, vol.~16,
  no.~6, pp. 769--786, 2012.

\bibitem{ED1}
K.~Price, R.~Storn, and J.~Lampinen, \emph{{Differential Evolution A Practical
  Approach to Global Optimization}}, ser. Natural Computing Series.\hskip 1em
  plus 0.5em minus 0.4em\relax Springer-Verlag, 2005. 

\bibitem{Mezura10a}
E.~Mezura-Montes, M.~E. Miranda-Varela, and R.~del Carmen G\'{o}mez-Ram\'{o}n,
  ``Differential evolution in constrained numerical optimization. an empirical
  study,'' \emph{Information Sciences}, vol. 180, no.~22, pp. 4223--4262, 2010.

\bibitem{RIGA}
J.~Grefenstette, ``Genetic algorithms for changing environments,'' in
  \emph{Parallel Problem Solving from Nature 2}.\hskip 1em plus 0.5em minus
  0.4em\relax Elsevier, 1992, pp. 137--144.

\bibitem{deb2000efficient}
K.~Deb, ``An efficient constraint handling method for genetic algorithms,''
  \emph{Computer methods in applied mechanics and engineering}, vol. 186,
  no.~2, pp. 311--338, 2000.

\bibitem{mezura2012empirical}
E.~Mezura-Montes and O.~Cetina-Dom{\'\i}nguez, ``Empirical analysis of a
  modified artificial bee colony for constrained numerical optimization,''
  \emph{Applied Mathematics and Computation}, vol. 218, no.~22, pp.
  10\,943--10\,973, 2012.

\bibitem{Derrac20113}
J.~Derrac, S.~Garc\'ia, D.~Molina, and F.~Herrera, ``A practical tutorial on
  the use of nonparametric statistical tests as a methodology for comparing
  evolutionary and swarm intelligence algorithms,'' \emph{Swarm and
  Evolutionary Computation}, vol.~1, no.~1, pp. 3--18, 2011. [Online].
  Available:
  \url{http://www.sciencedirect.com/science/article/pii/S2210650211000034}
\end{thebibliography}
\end{document}